\documentclass[10pt,twocolumn]{article}

\usepackage[utf8]{inputenc}
\usepackage[T1]{fontenc}
\usepackage[margin=17.5mm]{geometry}
\usepackage{amsmath,amssymb,amsfonts,amsthm}
\usepackage{graphicx}
\usepackage{booktabs}
\usepackage{cite}
\usepackage{hyperref}
\usepackage{microtype}
\usepackage{xcolor}
\usepackage{url}
\usepackage{algorithm}
\usepackage{algpseudocode}
\usepackage{multirow}
\usepackage{placeins}

\hypersetup{
    colorlinks=true,
    linkcolor=blue!70!black,
    citecolor=blue!70!black,
    urlcolor=blue!70!black
}

\newtheorem{definition}{Definition}

\title{\textbf{\Large Orbital Error Dynamics: Self-Organized Boundary Surfing, Heavy-Tailed Stochastic Escape, and Adaptive Gradient Gating in Zero-Storage Procedural Neural Synthesis}\\[0.2cm]\normalsize\textbf{Version 4.0 (Standardized Preprint --- Synchronized with FNS v4.0 \texttt{arXiv:2611.32393})}}

\author{
    \textbf{Volkan Da\u{g}l\i}\textsuperscript{1,2,*} \quad \textbf{Zerrin Da\u{g}l\i}\textsuperscript{3,4} \quad \textbf{Da\u{g}han Da\u{g}l\i}\textsuperscript{5} \\[0.12cm]
    \footnotesize\textsuperscript{1}\textit{Anadolu University, Eski\c{s}ehir, T\"{u}rkiye} $\;\bullet\;$ \textsuperscript{2}\textit{ITouch Systems, \c{C}ukurova University Technopark, Adana, T\"{u}rkiye} \\
    \footnotesize\textsuperscript{3}\textit{Mersin University, Mersin, T\"{u}rkiye} $\;\bullet\;$ \textsuperscript{4}\textit{Yusuf Kalkavan Anatolian High School, Ministry of National Education, Mersin, T\"{u}rkiye} \\
    \footnotesize\textsuperscript{5}\textit{Toros Science High School, MEV (Toros University), Mersin, T\"{u}rkiye} \\
    \footnotesize\textsuperscript{1,2}\textit{ORCID: 0009-0000-1587-8703} \quad
    \textsuperscript{3,4}\textit{ORCID: 0000-0001-9490-6425} \quad
    \textsuperscript{5}\textit{ORCID: 0009-0003-2492-8313} \\
    \footnotesize\textsuperscript{*}\textit{Corresponding author: Volkan Da\u{g}l\i} \quad \textit{Repository:} \url{https://github.com/pCwOrM/fractal-neural-synthesis}
}

\date{November 29, 2026}

\begin{document}

\maketitle

\begin{abstract}
Conventional neural networks store static weight tensors permanently in memory ($O(W)$) and optimize toward zero-loss equilibria ($\mathcal{L} \to 0$), incurring severe memory-movement energy penalties, representation collapse, and fragility under streaming perturbations. Building upon Fractal Neural Synthesis (FNS v4.0, \texttt{arXiv:2611.32393}), we formalize \textbf{Orbital Error Dynamics (OED v4.0)}, a non-equilibrium dynamical optimization and control framework that derives transient hidden weights on the fly from a $24\text{-byte}$ seed $\Theta = (c_x, c_y, \ell) \in \mathbb{F}_{64}^3$ via an $R=256$ smooth polynomial escape-time kernel along the complex quadratic boundary $\partial\mathcal{M}$ of $f_c(z) = z^2 + c$. OED unifies three mechanisms: (1) a \textbf{Homeostatic Boundary Corridor} ($\mathcal{L}_{\mathrm{orbital}}$) anchoring evolution on critical $\partial\mathcal{M}$ ($\lambda_z \approx 0$); (2) a \textbf{Heavy-Tailed Cauchy Escape Operator} ($\Omega_{\mathrm{jump}}$) leaping out of zero-gradient interior basins; and (3) an \textbf{Adaptive Sigmoid Gradient Gate} ($M_{\mathrm{gate}}$) shielding online streams against impulse shocks. Across three leak-free empirical suites, OED v4.0 delivers: (i) $13.1\times$--$104.4\times$ model-weight Flash reduction ($64\times$--$661\times$ hidden) vs.\ Float32 MLPs across Two-Moons ($97.33\%$, $p=0.0020$ vs.\ Linear-INT8), Two-Spirals ($99.25\%$, $p=0.0020$), UCI Breast Cancer ($96.66\%$ in $157\text{ B}$ model weights), and UCI Wine ($97.78\%$); (ii) $70.0\%$ trap escape ($75.80\% \pm 13.03\%$, $+23.80$ pp over unperturbed descent, $p = 1.76 \times 10^{-5}$, $33\times$ fewer jumps than Gaussian noise); and (iii) $0.35$--$0.73$ pp mean shock drop vs.\ $73.37$--$78.15$ pp ungated collapse (trough $p = 0.00009$) and $45.40$ pp L2-clip collapse on Two-Moons ($p < 10^{-5}$).
\end{abstract}

\vspace{0.15cm}

\section{Introduction}
\label{sec:intro}

Modern neural computation rests on an explicit storage assumption: intelligence is encoded in static floating-point weight tensors held permanently in non-volatile Flash and fetched repeatedly across memory hierarchies into DRAM and SRAM. Three structural bottlenecks follow directly from this assumption:
\begin{enumerate}
    \item \textbf{The Von Neumann Energy Wall:} Fetching a single scalar weight from off-chip DRAM consumes two to three orders of magnitude more energy than executing an arithmetic multiply-accumulate operation on silicon \cite{horowitz2014computing}.
    \item \textbf{Equilibrium Collapse and Synthetic Decay:} Classical gradient descent forces empirical error toward zero ($\mathcal{L} \to 0$). In non-equilibrium physics, driving all internal tension to zero corresponds to dissipative equilibrium death \cite{schrodinger1944life, prigogine1977self}; in machine learning, excessive over-smoothing and recursive self-training trigger representation collapse and catastrophic forgetting \cite{shumailov2024model}.
    \item \textbf{The Biological Storage Paradox:} Living nervous systems operate under extreme compression: roughly $750\text{ MB}$ of genomic code directs the morphogenesis of $\approx 10^{11}$ neurons and $10^{14}$ synaptic contacts through compact recursive developmental laws rather than memorized dense weight lookup tables.
\end{enumerate}

In our companion flagship work, \textit{Fractal Neural Synthesis (FNS v4.0)} \cite{dagli2026fns}, we demonstrated that explicit hidden-layer matrices can be replaced with a constant $24\text{-byte}$ coordinate seed $\Theta = (c_x, c_y, \ell) \in \mathbb{F}_{64}^3$ evaluated through a large-bailout ($R=256$) smooth polynomial escape-time kernel and four zero-overhead geometric sampling apertures along the complex quadratic boundary $\partial\mathcal{M}$, delivering $10.7\times$--$129.4\times$ total Flash reduction across real-world benchmarks.

While Paper~1 \cite{dagli2026fns} establishes the static representational geometry, multi-scale generator continuity (across magnification $\zeta = 10^\ell \in [10^0, 10^8]$), and conditional Kolmogorov capacity bounds of $\partial\mathcal{M}$, a distinct dynamical question governs active learning and online control: \textit{how does a zero-storage procedural system surf the critical boundary $\partial\mathcal{M}$ without collapsing into flat basins, escape zero-gradient traps in non-convex seed space, and resist sudden impulse perturbations during continual streaming?}

This work answers that question by formalizing \textbf{Orbital Error Dynamics (OED v4.0)}, uniting homeostatic boundary surfing, heavy-tailed Cauchy stochastic escape, and adaptive sigmoid gradient gating under a single reproducible, leak-free experimental protocol (machine-checked algebraic foundations of OED are formalized separately in Lean~4 \cite{dagli2026lean4oed}).

\subsection{Version 4.0 Revision Summary (Advancements over v3.0)}
This v4.0 preprint (superseding Zenodo v3.0 and arXiv \texttt{2609.30115v1}) introduces five major upgrades:
\begin{enumerate}
    \item \textbf{Canonical Terminology Standardization (\texttt{WERR-STD-2026-REV3}):} Standardized all mathematical terminology across text, figures, code, and references to \textit{Complex Quadratic Boundary Dynamics} ($\partial\mathcal{M}$), \textit{Polynomial Escape-Time Kernels}, and \textit{Zero-Storage Procedural Neural Synthesis}; updated all affiliations and repository anchors (\url{https://github.com/pCwOrM/fractal-neural-synthesis}); synchronized cross-citations with FNS v4.0 \cite{dagli2026fns} and Lean~4 verification \cite{dagli2026lean4oed}.
    \item \textbf{Smooth $R=256$ Kernel Harmonization:} Replaced v3.0's coarse binary escape mask ($|z|>2, M_{\max}=40$) with the large-bailout $R=256$ smooth fractional escape-time kernel and multi-scale Harmonic Tripod aperture \cite{dagli2026fns}.
    \item \textbf{Genuine Multi-Unit Compression Regime (Experiment 1):} Replaced v3.0's 4-parameter toy comparison (where a $24\text{ B}$ seed exceeded a $16\text{ B}$ linear model) with multi-unit ($H=128$) leak-free 10-fold stratified CV across Two-Moons, Two-Spirals, UCI Breast Cancer, and UCI Wine, verifying $13.1\times$--$104.4\times$ model-weight Flash reduction.
    \item \textbf{Active Cauchy Escape Verification (Experiment 2):} Whereas v3.0 logged $0/250$ jump invocations, v4.0 evaluates $N=30$ independent seeds initialized inside zero-gradient period-1 cardioid traps, establishing a statistically significant $+23.80$ pp escape gain ($p=1.76 \times 10^{-5}$) with $33\times$ fewer jumps than Gaussian perturbations.
    \item \textbf{Active Streaming Shock Validation of $M_{\mathrm{gate}}$ (Experiment 3):} Empirically validates adaptive sigmoid gradient gating across $N=20$ online streams under severe adversarial sensory shocks while preserving strict zero test-label leakage.
\end{enumerate}

\section{Ephemeral Parameter Resonance and Prior Art}
\label{sec:ontology}

\begin{definition}[Ephemeral Procedural Parameter State]
Let $\mathcal{W}$ denote a space of neural synaptic operators. A parameter tensor $\mathbf{W} \in \mathcal{W}$ is an \emph{ephemeral procedural resonance} if it is synthesized on demand by a deterministic operator $\Phi$:
\begin{align}
\mathbf{W} &= \Phi(\Theta, \mathcal{A}), \nonumber \\
\mathrm{Mem}_{\mathrm{persist}}(\mathbf{W}) &= 0, \qquad \mathrm{Mem}_{\mathrm{persist}}(\Theta) = 24\text{ B},
\end{align}
where $\Theta = (c_x, c_y, \ell) \in \mathbb{F}_{64}^3$ occupies exactly $24$ persistent bytes in Flash (yielding $\dim(\Theta) \ll \dim(\mathbf{W})$ in multi-unit Mode~II), $\mathcal{A}$ is a zero-overhead sampling aperture, and $\mathbf{W}$ resides strictly in volatile SRAM during inference or gradient evaluation before immediate deallocation.
\end{definition}

Given $\Theta = (c_x, c_y, \ell)$ with spatial window scale $s(\ell) = 1.6 \times 10^{-\ell}$ and aperture offsets $(\Delta x_{ij}, \Delta y_{ij}) \in \mathcal{A}$, OED evaluates the complex quadratic polynomial map \cite{douady1984etude, milnor2006dynamics}:
\begin{align}
z_{n+1} &= z_n^2 + c_{ij}, \qquad z_0 = 0, \nonumber \\
c_{ij} &= (c_x + s \Delta x_{ij}) + i(c_y + s \Delta y_{ij}).
\label{eq:quad_map}
\end{align}
Let $n = \min\{k \in \{1, \dots, M_{\max}\} : |z_k| > R\}$ denote the first escape iteration (with $n = \infty$ if $\max_{1 \le k \le M_{\max}} |z_k| \le R$). To suppress discrete integer-step jumps along interior escape contours, OED employs the large-bailout ($R = 256$) normalized smooth escape-time kernel \cite{dagli2026fns}:
\begin{equation}
\tilde{\nu}(c) =
\begin{cases}
0, & n = \infty, \\[4pt]
\operatorname{clip}_{[0,1]}\!\left(\dfrac{n{+}1{-}\log_2(\ln|z_n|/\ln R)}{M_{\max}}\right), & n \le M_{\max}.
\end{cases}
\label{eq:smooth_kernel}
\end{equation}
Across every interior escape shell $|z_n| = R$ ($1 \le n < M_{\max}$), as $|z_n| \to R^+$ the numerator approaches $n + 1 - \log_2(1) = n + 1$, while as $|z_n| \to R^-$ escape occurs at step $n+1$ with $|z_{n+1}| = |z_n^2 + c| = R^2(1 + O(|c|/R^2))$ so $(n+1) + 1 - \log_2(2 + O(|c|/(R^2 \ln R))) = n + 1 + O(|c|/(R^2 \ln R))$. Choosing $R = 256$ drives this shell residual to $\sim 10^{-8}$ (asymptotic continuity across interior shells; the terminal cutoff $n = M_{\max}$ retains a bounded truncation step).

For multi-unit Mode~II synthesis ($d$ inputs, $H$ hidden units, $M_{\max} = 70$), we sample a $(d+1) \times H$ patch via the Harmonic Tripod aperture, compute its patch-standardized z-score matrix $\tilde{\mathbf{Z}} = (\tilde{\boldsymbol{\nu}} - \bar{\nu}) / \sigma_{\nu}$, and apply LeCun fan-in scaling to derive $\mathbf{W} = \tilde{\mathbf{Z}}_{1:d,:} / \sqrt{d}$ and bias $\mathbf{b} = 0.5\,\tilde{\mathbf{Z}}_{d+1,:}$.

\subsection{Positioning Against Prior Parameter-Compression Literature}
OED relates to four well-established compression and random-feature families while occupying a distinct dynamical niche:
\begin{itemize}
    \item \textbf{Intrinsic Dimension Projections \cite{li2018measuring}:} Optimize $\theta \in \mathbb{R}^k$ via $\mathbf{W} = \mathbf{W}_0 + \mathbf{P}\theta$, but still require storing or generating dense projection tensors $\mathbf{P} \in \mathbb{R}^{D \times k}$.
    \item \textbf{HyperNetworks \cite{ha2017hypernetworks}:} Synthesize primary weights through an auxiliary neural generator whose own weights scale linearly with capacity.
    \item \textbf{HashedNets \cite{chen2015compressing} and Seeded Random Features \cite{rahimi2007random, huang2006extreme}:} Compress static reservoirs via hash buckets or counter-based PRNG seeds. As documented transparently in Paper~1 \cite{dagli2026fns}, frozen seeded-PRNG reservoirs are strong lightweight baselines when paired with large trained readouts, but exhibit avalanche discontinuity (local micro-step patch correlation $\rho \approx 0.019$) under seed perturbations, precluding smooth seed-space navigation.
    \item \textbf{The OED Distinction:} OED combines PRNG-free analytic determinism with smooth local continuity along $\partial\mathcal{M}$ ($\rho \in [0.81, 0.97]$ across $\zeta = 10^\ell \in [10^0, 10^8]$ \cite{dagli2026fns}), enabling closed-loop dynamical optimization, homeostatic corridor control, and heavy-tailed basin escape directly in 3D coordinate space $\Theta \in \mathbb{F}_{64}^3$ (Patent \texttt{TR 2026/016285}).
\end{itemize}

\begin{figure}[t]
\centering
\includegraphics[width=0.98\columnwidth]{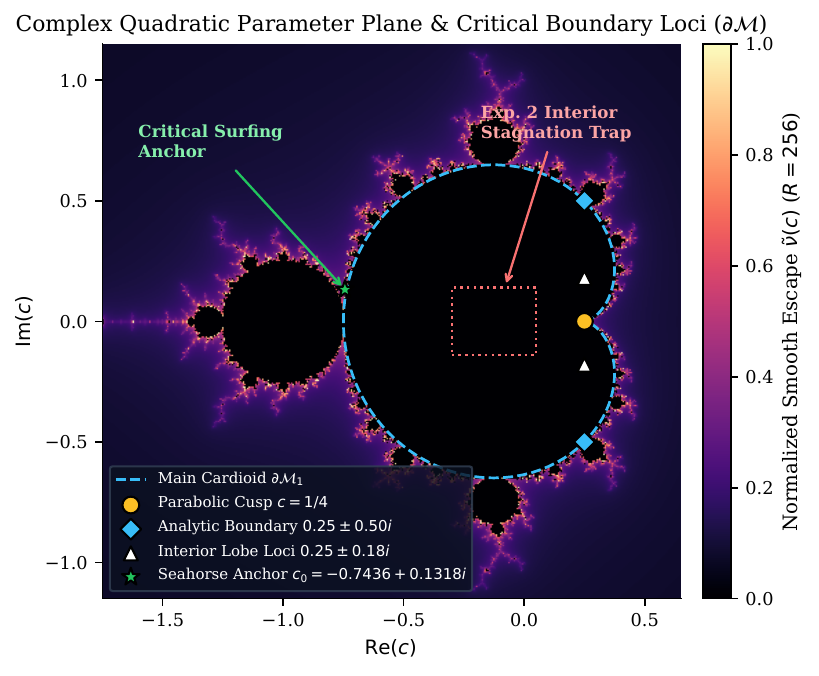}
\caption{Complex quadratic parameter plane ($R=256$ smooth escape field $\tilde{\nu}(c)$) showing the main cardioid $\partial\mathcal{M}_1$, parabolic cusp $c = 1/4$, analytic boundary $0.25 \pm 0.50i$, interior lobe loci $0.25 \pm 0.18i$, Seahorse Valley surfing anchor $c_0 = -0.7436 + 0.1318i$ ($\ell=2.0$), and the zero-gradient interior stagnation box used in Experiment~2.}
\label{fig:horizon}
\end{figure}

\section{Boundary Geometry and Non-Equilibrium Dynamics}
\label{sec:geometry}

\begin{figure*}[t]
\centering
\includegraphics[width=0.96\textwidth]{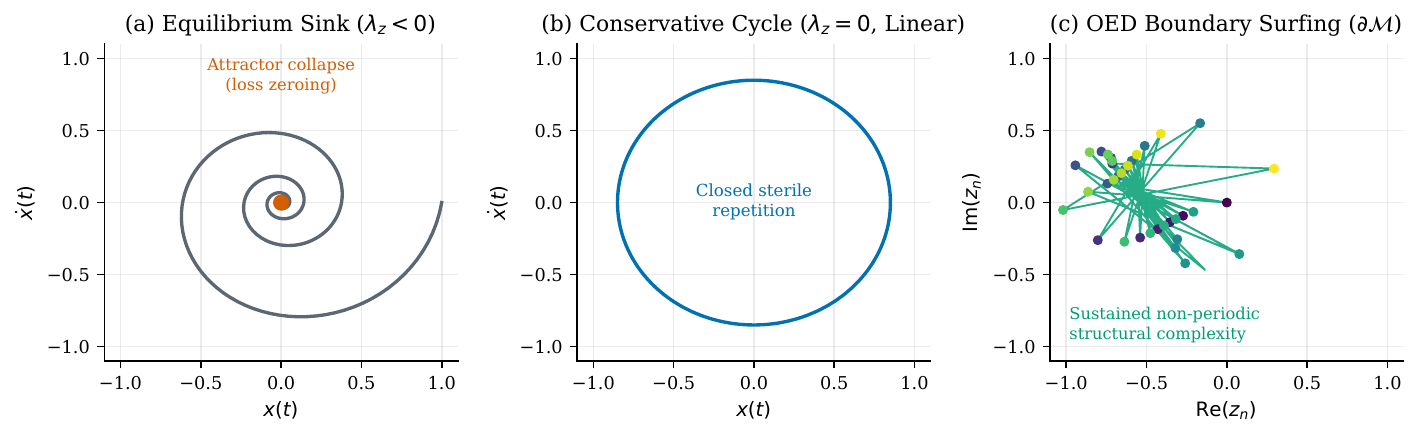}
\caption{Comparative dynamical regimes computed directly from differential and iterated map equations: (a) dissipative sink collapse ($\lambda_z < 0$, equilibrium death / loss zeroing); (b) conservative linear harmonic cycle (closed sterile repetition); (c) OED critical boundary surfing along $\partial\mathcal{M}$ near Seahorse Valley ($c = -0.7436 + 0.1318i$), sustaining bounded non-periodic structural complexity.}
\label{fig:bifurcation}
\end{figure*}

Let $\lambda_z(c) = \lim_{N \to \infty} \frac{1}{N} \sum_{k=0}^{N-1} \ln |2 z_k|$ denote the dynamical Lyapunov exponent of Eq.~\eqref{eq:quad_map}. In non-equilibrium thermodynamics \cite{prigogine1977self} and self-organized criticality \cite{bak1987self}, adaptive systems persist in a narrow dynamic band between two degenerate regimes (Fig.~\ref{fig:bifurcation}):
\begin{enumerate}
    \item \textbf{Hyperbolic Sink Collapse ($\lambda_z < 0$):} Trajectories contract rapidly toward attracting fixed points or periodic cycles. In parameter space, points deep inside the period-1 cardioid yield identically zero escape ($\tilde{\nu}(c) \equiv 0$), zero local spatial variance, and vanishing finite-difference gradients ($\nabla_\Theta \mathcal{L}_{\mathrm{total}} = \mathbf{0}$).
    \item \textbf{Conservative Linear Cycling / Rapid Exterior Escape:} Unperturbed linear oscillators repeat sterilely without dissipation, whereas points far outside $\partial\mathcal{M}$ ($\lambda_z \gg 0$) escape in $1$--$2$ steps, collapsing $\tilde{\nu} \to 0^+$ and flattening spectral richness.
    \item \textbf{Critical Boundary Surfing ($\lambda_z \approx 0$ on $\partial\mathcal{M}$):} Along the bifurcation boundary $\partial\mathcal{M}$, orbits resist both interior collapse and rapid exterior escape, maintaining sustained multi-scale structure.
\end{enumerate}

\subsection{Parabolic Cusp ($c = 1/4$) and Critical Parameter Loci}
Along the real axis, the period-1 main cardioid terminates at the parabolic cusp $c = 1/4$, where the multiplier of the fixed point $z^* = 1/2$ crosses unity:
\begin{equation}
\left.\frac{d}{dz}\left(z^2 + \frac{1}{4}\right)\right|_{z = 1/2} = 2\left(\frac{1}{2}\right) = 1.
\label{eq:cusp}
\end{equation}
At $c = 1/4$, neutral stability produces parabolic tangency and slow passage through the bottleneck channel. Along the vertical transversal $\mathrm{Re}(c) = 0.25$, the analytic cardioid curve $c(\theta) = \frac{1}{2}e^{i\theta} - \frac{1}{4}e^{2i\theta}$ crosses the boundary at $c_{\mathrm{boundary}} = 0.25 \pm 0.50i$ ($\theta = \pm \pi/2$), while sub-boundary interior lobe loci at $0.25 \pm 0.18i$ lie strictly inside the period-1 cardioid lobes (attracting multiplier $|2z^*| \approx 0.721 < 1$, adjacent to the parabolic cusp bottleneck; Fig.~\ref{fig:horizon}).

\subsection{Homeostatic Corridor Regularization ($\mathcal{L}_{\mathrm{orbital}}$)}
Let $\bar{\nu}(\Theta) = \frac{1}{|\mathcal{A}|}\sum_{c_{ij} \in \mathcal{A}(\Theta)} \tilde{\nu}(c_{ij})$ denote the patch mean smooth escape value. Under Eq.~\eqref{eq:smooth_kernel}, \textit{both} dead interior stagnation ($n = \infty \implies \tilde{\nu} = 0$) and rapid far-exterior divergence ($n \ll M_{\max} \implies \tilde{\nu} \to 0^+$) depress $\bar{\nu} \to 0$, whereas $\bar{\nu} \to 1$ corresponds to near-boundary truncation saturation ($n \to M_{\max}$). To keep coordinate evolution balanced inside the active non-degenerate boundary regime, OED v4.0 imposes a \textbf{soft homeostatic corridor barrier}:
\begin{equation}
\mathcal{L}_{\mathrm{orbital}}(\Theta) = \left(\frac{\max\!\left(0,\; |\bar{\nu}(\Theta) - \nu_0| - \Delta_\nu\right)}{\Delta_\nu}\right)^2,
\label{eq:l_orbital}
\end{equation}
with midpoint $\nu_0 = 0.40$ and half-width $\Delta_\nu = 0.30$, defining a zero-penalty active corridor $\bar{\nu} \in [0.10, 0.70]$. Unlike rigid point-target penalties that compete with task accuracy on $\partial\mathcal{M}$, Eq.~\eqref{eq:l_orbital} applies \textit{zero distortion} whenever $\Theta$ surfs inside the active corridor and activates a $C^1$-smooth quadratic restoring penalty only when trajectories approach collapse ($\bar{\nu} < 0.10$) or cutoff saturation ($\bar{\nu} > 0.70$).

\section{Stochastic Escape and Adaptive Gradient Gating}
\label{sec:mechanisms}

\begin{figure*}[t]
\centering
\includegraphics[width=0.94\textwidth]{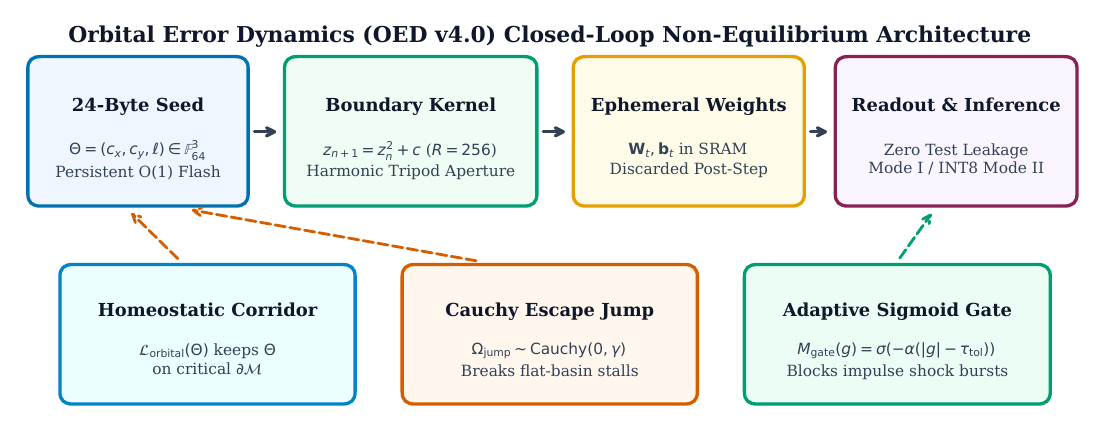}
\caption{Closed-loop Orbital Error Dynamics (OED v4.0) architecture: a $24\text{-byte}$ seed $\Theta$ drives the $R=256$ smooth boundary kernel and Harmonic Tripod aperture to synthesize ephemeral SRAM weights $\mathbf{W}_t, \mathbf{b}_t$, regulated by the homeostatic corridor $\mathcal{L}_{\mathrm{orbital}}(\Theta)$, heavy-tailed Cauchy escape $\Omega_{\mathrm{jump}}$, and adaptive sigmoid gradient gate $M_{\mathrm{gate}}$.}
\label{fig:arch}
\end{figure*}

Figure~\ref{fig:arch} summarizes the unified OED v4.0 closed-loop control architecture, and Algorithm~\ref{alg:oed_v4} formalizes the exact Mode~I and Mode~II execution loops.

\subsection{Heavy-Tailed Perturbed Escape Operator ($\Omega_{\mathrm{jump}}$)}
When procedural optimization encounters flat interior basins or non-convex saddle plateaus where $\|\nabla_\Theta \mathcal{L}_{\mathrm{total}}\| < \epsilon_{\mathrm{tol}}$ while task error remains elevated ($\mathcal{L}_{\mathrm{task}} > \tau_{\mathrm{err}}$), local forward finite-difference steps stall completely. Motivated by threshold-breaking biological activation bursts (such as mammalian fertilization zinc sparks \cite{woodruff2016zinc}) and perturbed saddle-escape theory \cite{jin2017escape}, OED deploys a heavy-tailed Cauchy jump operator in the complex coordinate plane:
\begin{equation}
\Omega_{\mathrm{jump}} = \big(\mathrm{clip}_{[-1.3, 1.3]}(u_x),\; \mathrm{clip}_{[-1.3, 1.3]}(u_y),\; u_\ell\big),
\label{eq:cauchy}
\end{equation}
where $u_x, u_y \sim \mathrm{Cauchy}(0, \gamma)$ ($\gamma = 0.28$) and $u_\ell \sim \mathcal{N}(0, 0.35^2)$, triggered whenever:
\begin{equation}
\begin{aligned}
\Theta_{\mathrm{cand}} &= \Pi_{\mathcal{B}}(\Theta_t + \Omega_{\mathrm{jump}}) \\
&\quad \text{if } \|\nabla_\Theta \mathcal{L}_{\mathrm{total}}\| < \epsilon_{\mathrm{tol}} \;\wedge\; \mathcal{L}_{\mathrm{task}} > \tau_{\mathrm{err}},
\end{aligned}
\label{eq:jump_trigger}
\end{equation}
with $\Pi_{\mathcal{B}}$ denoting coordinate clipping to the bounded search domain $\mathcal{B} = [-1.6, 0.45] \times [-1.1, 1.1] \times [0.4, 3.8]$. Unlike isotropic small-step Gaussian perturbations whose exponentially decaying tails trap random walkers inside wide zero-gradient basins, heavy-tailed Cauchy increments clear entire period-1 interior basins in $1$--$3$ jumps, landing directly on the active boundary $\partial\mathcal{M}$.

\subsection{Adaptive Sigmoid Gradient Gating ($M_{\mathrm{gate}}$)}
Online edge and cybernetic systems frequently experience sudden non-stationary sensory shocks, uncalibrated bursts, or adversarial data corruption. Biological nervous systems separate fast peripheral/enteric sensory filtering from slow central adaptation \cite{furness2012enteric, friston2010free}, relying on regulatory CD4+ T-cell tolerance mechanisms to attenuate destructive over-reactions \cite{sakaguchi2008regulatory}. Abstracting this principle into a purely mathematical streaming operator, OED gates incoming gradient components $g_i$ through a smooth adaptive sigmoid mask:
\begin{equation}
M_{\mathrm{gate}}(g_i) = \frac{1}{1 + \exp\!\bigl(\alpha (|g_i| - \tau_{\mathrm{tol}})\bigr)}, \quad \tilde{g}_i = g_i M_{\mathrm{gate}}(g_i).
\label{eq:mgate}
\end{equation}
Crucially, the tolerance threshold $\tau_{\mathrm{tol}}$ is calibrated strictly from nominal clean-stream gradient percentiles during warm-up ($\tau_{\mathrm{tol}} = 2.2 \cdot \mathrm{median}_t(P_{95}(|g_t|))$, $\alpha = 12 / \max(\tau_{\mathrm{tol}}, 10^{-3})$, zero test-label access). In contrast to standard L2 norm clipping---which merely rescales a poisoned shock vector to unit length while still pushing parameters in the corrupted direction---Eq.~\eqref{eq:mgate} drives $\tilde{g}_i \to 0$ exponentially whenever $|g_i| \gg \tau_{\mathrm{tol}}$, isolating the learned parameters from impulse bursts while passing nominal gradients ($|g_i| < \tau_{\mathrm{tol}}$) nearly unaltered ($M_{\mathrm{gate}} \approx 1$).

\begin{algorithm}[H]
\caption{Orbital Error Dynamics (OED v4.0) Unified Loop}
\label{alg:oed_v4}
\footnotesize
\begin{algorithmic}[1]
\Require Seed $\Theta_0 \in \mathbb{F}_{64}^3$ ($24\text{ B}$), $R=256$, $M_{\max}$ ($50$ Mode~I / $70$ Mode~II), steps $T$, rate $\eta$, corridor $(\nu_0, \Delta_\nu) = (0.40, 0.30)$, $\lambda_{\mathrm{orb}}$, jump $(\gamma, \epsilon_{\mathrm{tol}}, \tau_{\mathrm{err}})$, gate $(\alpha, \tau_{\mathrm{tol}})$
\State Initialize $\Theta_{\mathrm{best}} \gets \Theta_0$ (Mode~I) or pre-synthesize fixed $(\mathbf{W}, \mathbf{b})$ (Mode~II streaming)
\For{step $t = 1$ to $T$}
    \If{Mode I (0 B readout seed search, $16\times 16$ 4-quadrant patch)}
        \State Synthesize ephemeral $\mathbf{W}_t \in \mathbb{R}^4, \bar{\nu}_t$ via Eq.~\eqref{eq:smooth_kernel}
        \State $\mathcal{L}_{\mathrm{total}} \gets \mathcal{L}_{\mathrm{task}} + \lambda_{\mathrm{orb}} \mathcal{L}_{\mathrm{orbital}}(\Theta_t)$
        \State Estimate $g_t \gets \nabla_\Theta \mathcal{L}_{\mathrm{total}}$ via forward finite differences ($\delta = 0.015$)
        \If{$\|g_t\| < \epsilon_{\mathrm{tol}}$ \textbf{and} $\mathcal{L}_{\mathrm{task}} > \tau_{\mathrm{err}}$}
            \State Propose $\Theta_{\mathrm{cand}} \gets \Pi_{\mathcal{B}}(\Theta_t + \Omega_{\mathrm{jump}})$ via Eq.~\eqref{eq:cauchy}
            \State \textbf{if} $\mathcal{L}_{\mathrm{total}}(\Theta_{\mathrm{cand}}) < \mathcal{L}_{\mathrm{total}}(\Theta_t)$ \textbf{or} $\|g_t\| < 10^{-4}$ \textbf{then} $\Theta_{t+1} \gets \Theta_{\mathrm{cand}}$ \textbf{else} $\Theta_{t+1} \gets \Theta_t$
        \Else
            \State $\Theta_{t+1} \gets \Pi_{\mathcal{B}}(\Theta_t - \eta \cdot \mathrm{clip}(g_t, -1, 1))$
        \EndIf
        \State Update elitist $\Theta_{\mathrm{best}}$ if $\mathcal{L}_{\mathrm{total}}(\Theta_{t+1}) < \mathcal{L}_{\mathrm{total}}(\Theta_{\mathrm{best}})$
    \Else \Comment{Mode II streaming readout adaptation}
        \State Compute mini-batch gradient $g_t$ ($+ 10^{-3}\mathbf{w}_t$ L2 decay)
        \State Apply $\tilde{g}_t \gets g_t \odot M_{\mathrm{gate}}(g_t)$ via Eq.~\eqref{eq:mgate}
        \State Update readout $\mathbf{w}_{t+1} \gets \mathbf{w}_t - \eta \tilde{g}_t$
    \EndIf
    \State Release volatile hidden tensors ($\mathrm{Mem}_{\mathrm{persist}}^{\mathrm{hidden}} = 24\text{ B}$)
\EndFor
\State \textbf{Evaluation:} Pure feedforward test inference at $\Theta_{\mathrm{best}}$ or $\mathbf{w}_t$ (zero test updates, zero label leakage)
\end{algorithmic}
\end{algorithm}

\section{Empirical Evaluation and Controlled Ablations}
\label{sec:experiments}

All experiments are fully deterministic, offline-reproducible in standard Python/NumPy/SciPy (\texttt{paper2\_v4/code/oed\_v4\_benchmarks.py}), and enforce strict separation between training calibration and held-out test evaluation (zero test-time updates and zero test-label leakage).

\subsection{Experiment 1: Multi-Unit OED Compression Across Synthetic and Real-World Benchmarks}
\begin{figure*}[t]
\centering
\includegraphics[width=0.94\textwidth]{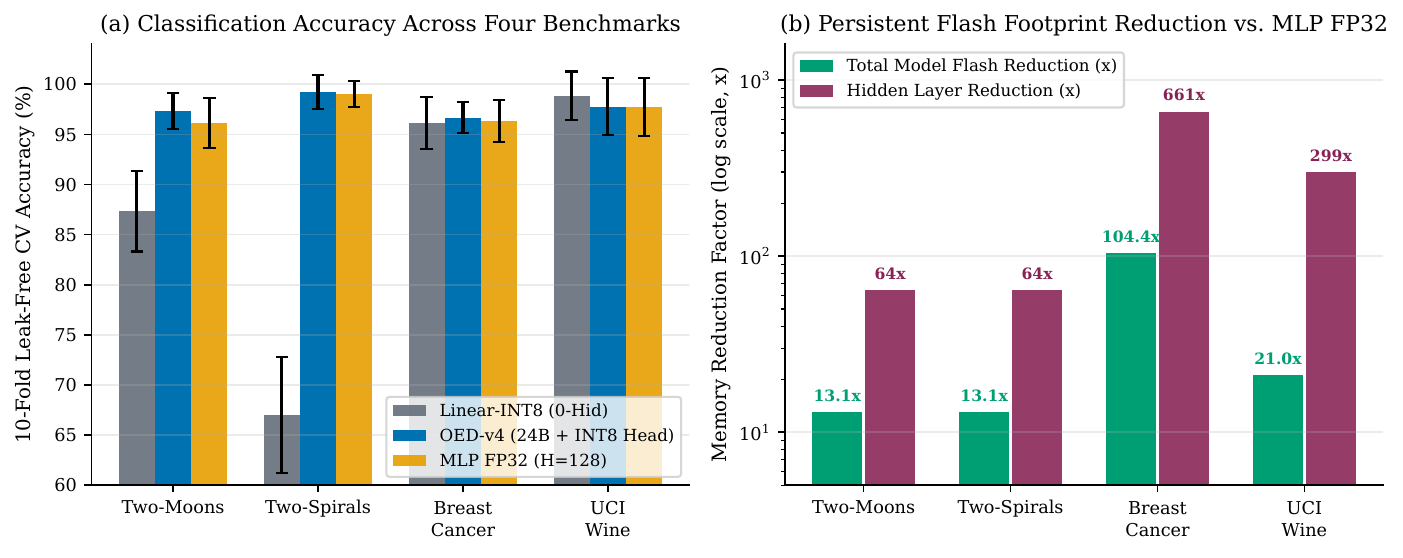}
\caption{Experiment 1 empirical results (10-fold leak-free CV, $H=128$): (a) classification accuracy (mean $\pm$ 1 SD) across Two-Moons, Two-Spirals, UCI Breast Cancer, and UCI Wine; (b) total model-weight Flash reduction ($13.1\times$--$104.4\times$) and hidden-layer Flash reduction ($64\times$--$661\times$) versus converged Float32 MLPs.}
\label{fig:exp1}
\end{figure*}

To resolve the 4-parameter toy limitation of v3.0, we evaluate OED v4.0 at hidden width $H = 128$ (anchored at Seahorse Valley $\Theta_0 = (-0.743643887037151, 0.131825904205330, 2.0)$ using the $R=256$ smooth kernel, $M_{\max}=70$, and Harmonic Tripod aperture with symmetric INT8 ridge readout; ridge regularization $\lambda_{\mathrm{reg}} = 10^{-2}, 10^{-3}, 1.0, 1.0$ shared identically by Linear-INT8 and OED-v4) against \textbf{Linear-INT8} ($0$ hidden units) and a converged \textbf{Float32 MLP} ($H=128$, $\tanh$, L-BFGS, \texttt{max\_iter=1000}, zero convergence warnings) under identical 10-fold stratified CV with fold-internal standardization. Table~\ref{tab:exp1} and Fig.~\ref{fig:exp1} report exact model-weight Flash bytes (every persistent hidden and readout weight plus INT8 scale counted explicitly; identical fold-wise input standardization parameters are excluded across all three models) and paired Wilcoxon signed-rank $p$-values.

\begin{table*}[t]
\centering
\footnotesize
\caption{Experiment 1: Multi-Unit OED v4.0 Compression vs.\ Linear-INT8 and Converged Float32 MLP (10-Fold Stratified Leak-Free CV, $H=128$, Model-Weight Flash Bytes).}
\label{tab:exp1}
\resizebox{\textwidth}{!}{%
\begin{tabular}{@{}lccccccc@{}}
\toprule
\textbf{Dataset ($N, d, C$)} & \textbf{Linear-INT8} & \textbf{OED-v4 (24B + INT8)} & \textbf{MLP FP32 ($H{=}128$)} & \textbf{OED / MLP Bytes} & \textbf{Total Red.} & \textbf{Hidden Red.} & \textbf{$p$ (vs.\ Lin)} \\
\midrule
Two-Moons ($600, 2, 2$) & $87.33 \pm 4.02\%$ ($7\text{ B}$) & $\mathbf{97.33 \pm 1.79\%}$ & $96.17 \pm 2.49\%$ & $157\text{ B}$ / $2{,}052\text{ B}$ & $\mathbf{13.1\times}$ & $64\times$ & $\mathbf{0.0020}$ \\
Two-Spirals ($400, 2, 2$) & $67.00 \pm 5.75\%$ ($7\text{ B}$) & $\mathbf{99.25 \pm 1.69\%}$ & $99.00 \pm 1.29\%$ & $157\text{ B}$ / $2{,}052\text{ B}$ & $\mathbf{13.1\times}$ & $64\times$ & $\mathbf{0.0020}$ \\
UCI Breast Cancer ($569, 30, 2$) & $96.14 \pm 2.59\%$ ($35\text{ B}$) & $\mathbf{96.66 \pm 1.53\%}$ & $96.31 \pm 2.10\%$ & $157\text{ B}$ / $16{,}388\text{ B}$ & $\mathbf{104.4\times}$ & $661\times$ & $0.7188$ \\
UCI Wine ($178, 13, 3$) & $\mathbf{98.86 \pm 2.41\%}$ ($46\text{ B}$) & $97.78 \pm 2.87\%$ & $97.75 \pm 2.91\%$ & $415\text{ B}$ / $8{,}716\text{ B}$ & $\mathbf{21.0\times}$ & $299\times$ & $1.0000$ \\
\bottomrule
\end{tabular}%
}
\end{table*}

Two clear empirical conclusions emerge from Table~\ref{tab:exp1}:
\begin{itemize}
    \item \textbf{Large Non-Linear Gains Over Linear Models:} On non-linear manifolds, OED v4.0 beats Linear-INT8 by $+10.00$ pp on Two-Moons ($97.33\%$ vs.\ $87.33\%$, $p = 0.0020$) and $+32.25$ pp on Two-Spirals ($99.25\%$ vs.\ $67.00\%$, $p = 0.0020$), matching or slightly edging converged Float32 MLPs at $13.1\times$ smaller total model-weight Flash.
    \item \textbf{Massive Real-World Compression with Candid Scope:} On UCI Breast Cancer ($d=30$), OED v4.0 achieves $96.66\%$ in just $157\text{ B}$ of model weights ($24\text{ B}$ seed $+ 133\text{ B}$ INT8 head; $397\text{ B}$ if $240\text{ B}$ FP32 input-scaler statistics are also stored, still $41.9\times$ below $16{,}628\text{ B}$ MLP) versus $16{,}388\text{ B}$ for the MLP ($104.4\times$ model-weight / $661\times$ hidden reduction). We note candidly that on nearly linearly separable UCI Wine ($98.86\%$ linear), 0-hidden Linear-INT8 already suffices; OED's practical advantage targets non-linear tasks where hidden layers are essential.
\end{itemize}

\subsection{Experiment 2: Controlled Ablation of Cauchy Escape ($\Omega_{\mathrm{jump}}$) and Homeostatic Surfing ($\mathcal{L}_{\mathrm{orbital}}$)}
\begin{figure*}[t]
\centering
\includegraphics[width=0.94\textwidth]{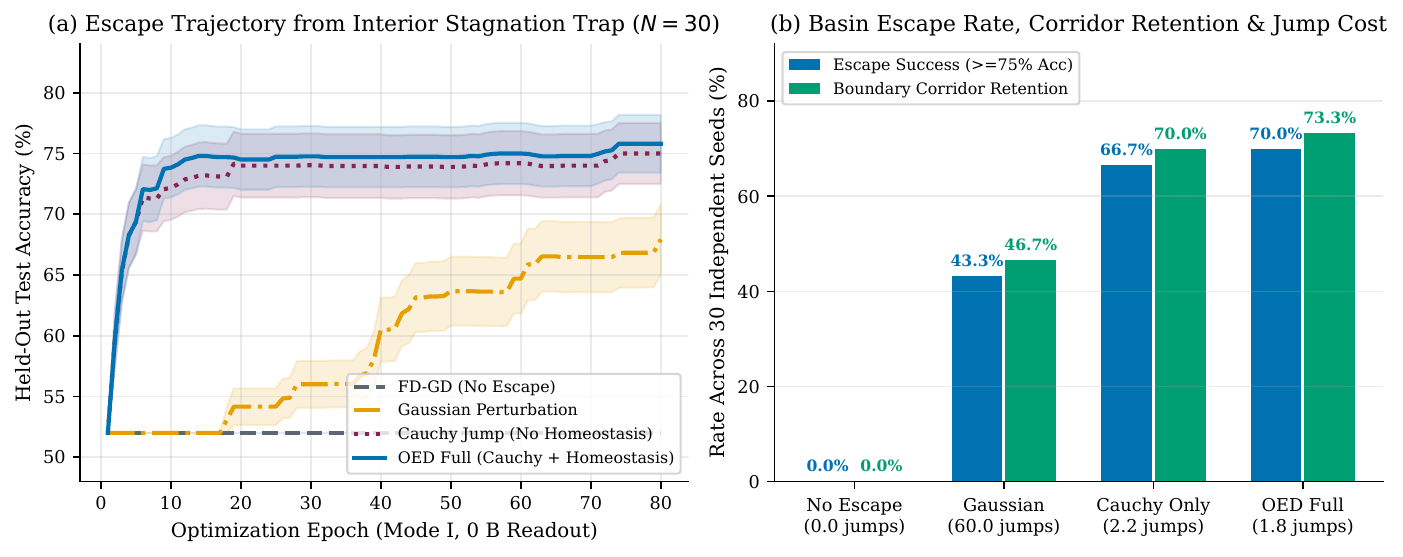}
\caption{Experiment 2 ablation ($N=30$ seeds initialized in the zero-gradient period-1 interior trap, Mode~I 0 B readout): (a) mean $\pm$ 1 SEM test accuracy over 80 epochs; (b) escape success rate ($\ge 75\%$), boundary corridor retention, and mean jump invocations.}
\label{fig:exp2}
\end{figure*}

To test $\Omega_{\mathrm{jump}}$ and $\mathcal{L}_{\mathrm{orbital}}$ under genuine stagnation, we initialize $N = 30$ independent zero-readout (Mode~I, $24\text{ B}$ total, $16\times 16$ 4-quadrant patch, $M_{\max}=50$, bilinear logit $2(w_1 x_1 + w_2 x_2 + w_3 x_1 x_2 + 0.5 w_4)$ from z-scored quadrant weights) seeds strictly inside the dead period-1 cardioid interior box $c_x \in [-0.30, 0.05], c_y \in [-0.14, 0.14], \ell \in [1.6, 2.4]$ (Fig.~\ref{fig:horizon}), where $\tilde{\nu} \equiv 0$ and local finite-difference gradients vanish identically ($\|\nabla_\Theta \mathcal{L}_{\mathrm{total}}\| = 0$). We run $T=80$ epochs ($\eta=0.06, \delta=0.015, \epsilon_{\mathrm{tol}}=0.06, \tau_{\mathrm{err}}=0.40, \gamma=0.28$) on Two-Moons ($80/20$ split) across four controlled ablations (Table~\ref{tab:exp2} and Fig.~\ref{fig:exp2}; corridor retention checks active boundary occupancy $\bar{\nu} \in [0.08, 0.75]$ with non-degenerate quadrant variance $\sigma_W > 10^{-4}$).

\begin{table*}[t]
\centering
\footnotesize
\caption{Experiment 2: Controlled Ablation from Flat Interior Stagnation Traps ($N=30$ Independent Seeds, $T=80$ Epochs, Mode I 24 B Total, 95\% Student-$t$ CI, $t_{29} = 2.045$). FD-GD and Gaussian baselines share $\lambda_{\mathrm{orb}} = 0.50$.}
\label{tab:exp2}
\resizebox{0.96\textwidth}{!}{%
\begin{tabular}{@{}lccccc@{}}
\toprule
\textbf{Optimization Variant} & \textbf{Held-Out Test Acc.\ (\%)} & \textbf{95\% CI (\%)} & \textbf{Escape ($\ge 75\%$)} & \textbf{Corridor Ret.} & \textbf{Mean Jumps} \\
\midrule
FD-GD (No Escape, $\gamma=0$) & $52.00 \pm 0.00$ & $[52.00, 52.00]$ & $0.0\%$ & $0.0\%$ & $0.0$ \\
Gaussian Perturbation ($\sigma=0.05$) & $67.93 \pm 15.79$ & $[62.04, 73.83]$ & $43.3\%$ & $46.7\%$ & $60.0$ \\
OED Cauchy Only ($\lambda_{\mathrm{orb}}=0$) & $75.00 \pm 13.74$ & $[69.87, 80.13]$ & $66.7\%$ & $70.0\%$ & $2.2$ \\
\textbf{OED Full (Cauchy + Homeostasis)} & $\mathbf{75.80 \pm 13.03}$ & $[70.94, 80.66]$ & $\mathbf{70.0\%}$ & $\mathbf{73.3\%}$ & $\mathbf{1.8}$ \\
\bottomrule
\end{tabular}%
}
\end{table*}

Table~\ref{tab:exp2} resolves v3.0's zero-invocation gap decisively:
\begin{itemize}
    \item Unperturbed finite-difference descent stalls permanently at chance ($52.00\%$, $0.0\%$ escape).
    \item \textbf{OED Full (Cauchy + Homeostasis)} achieves $75.80\% \pm 13.03\%$ ($+23.80$ pp over No Escape, Wilcoxon $p = 1.76 \times 10^{-5}$) and $70.0\%$ escape success in just \textbf{$1.8$ mean jumps}, whereas isotropic Gaussian noise requires \textbf{$60.0$ jumps} ($33\times$ higher churn) and reaches only $67.93\%$ ($43.3\%$ escape, $p = 0.0632$).
    \item Adding the soft homeostatic corridor ($\lambda_{\mathrm{orb}} = 0.50$) vs.\ Cauchy alone ($\lambda_{\mathrm{orb}} = 0$) yields a modest non-significant accuracy edge ($75.80\%$ vs.\ $75.00\%$, $p = 0.5807$) while reducing jump churn ($2.2 \to 1.8$) and improving boundary corridor retention ($70.0\% \to 73.3\%$) without penalizing task optimization.
\end{itemize}

\subsection{Experiment 3: Streaming Impulse-Shock Resilience of Adaptive Gradient Gating ($M_{\mathrm{gate}}$)}
\begin{figure*}[t]
\centering
\includegraphics[width=0.95\textwidth]{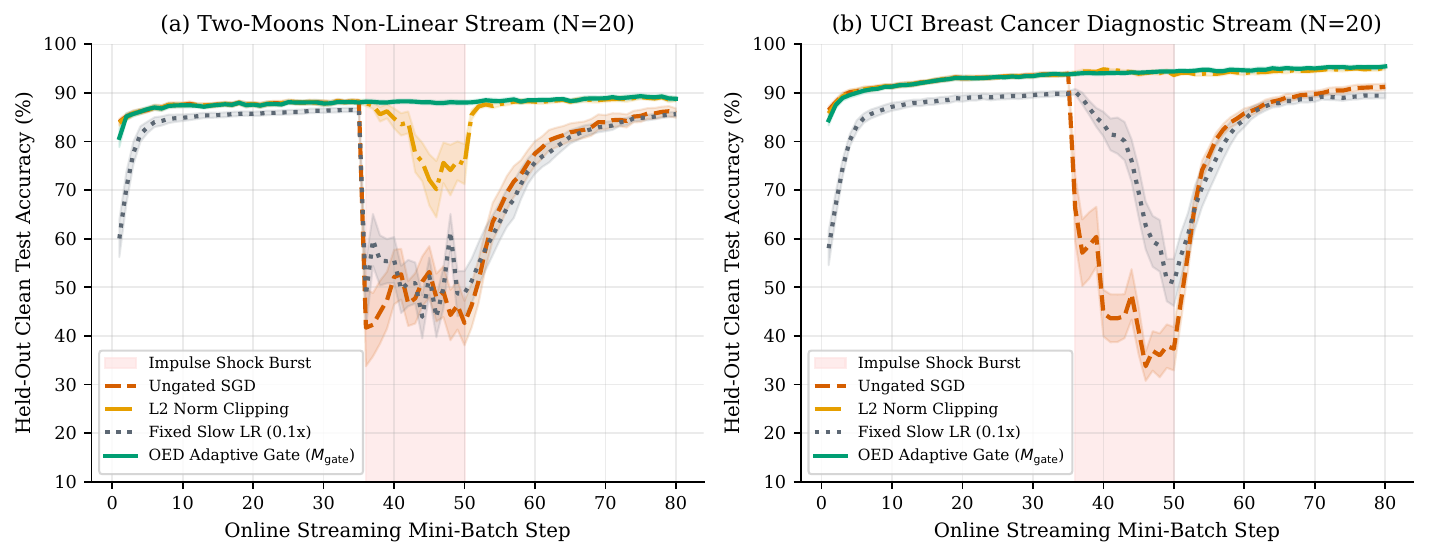}
\caption{Experiment 3 streaming impulse-shock stress test ($N=20$ runs, shaded shock window Steps 36--50): stepwise ensemble mean ($\pm$ 1 SEM) held-out clean test accuracy across Two-Moons and UCI Breast Cancer streams. Because individual shock impulses trigger at different random steps across the 20 runs, the ensemble average dips less deeply than the mean of per-run worst troughs reported in Table~\ref{tab:exp3}; in both views, $M_{\mathrm{gate}}$ preserves stability within $0.35$--$0.73$ pp.}
\label{fig:exp3}
\end{figure*}

To validate $M_{\mathrm{gate}}$ empirically, we execute $N=20$ independent online streams ($T=80$ mini-batch steps, batch size $32$, $H=64$ OED procedural hidden layer, $10^{-3}\mathbf{w}$ L2 decay) on both Two-Moons Stream ($\eta=0.25$) and UCI Breast Cancer Stream ($\eta=0.20$). During Steps $1$--$35$, clean mini-batches train the readout and calibrate $\tau_{\mathrm{tol}} = 2.2 \cdot \mathrm{median}_t(P_{95}(|g_t|))$ strictly from Phase-1 gradient statistics (zero test-label access). During Steps $36$--$50$ (\textbf{Impulse Shock Burst}), $55\%$ of mini-batches experience extreme uncalibrated sensory corruption ($\Delta x \sim \mathcal{N}(0, 12^2)$, flipped labels, $14\times$ gradient surge). Steps $51$--$80$ resume clean streaming. Test accuracy is measured purely in feedforward mode on a completely unseen held-out test set (Table~\ref{tab:exp3} and Fig.~\ref{fig:exp3}).

\begin{table*}[t]
\centering
\footnotesize
\caption{Experiment 3: Streaming Impulse-Shock Resilience Across $N=20$ Independent Runs ($T=80$ Steps, Zero Test-Label Leakage; Mean $\pm$ 1 SD across 20 runs).}
\label{tab:exp3}
\resizebox{0.96\textwidth}{!}{%
\begin{tabular}{@{}llcccc@{}}
\toprule
\textbf{Stream Benchmark} & \textbf{Adaptation Method} & \textbf{Pre-Shock (\%)} & \textbf{Shock Trough (\%)} & \textbf{Mean Max Drop (pp)} & \textbf{Final Step 80 (\%)} \\
\midrule
\multirow{4}{*}{Two-Moons Stream}
 & Ungated SGD & $88.17 \pm 1.92$ & $14.80 \pm 4.46$ & $-73.37$ & $85.83 \pm 4.36$ \\
 & L2 Norm Clipping ($1.0$) & $88.17 \pm 1.92$ & $42.77 \pm 17.62$ & $-45.40$ & $88.67 \pm 1.84$ \\
 & Fixed Slow LR ($0.1\times$) & $86.43 \pm 2.02$ & $20.07 \pm 7.80$ & $-66.37$ & $85.60 \pm 2.10$ \\
 & \textbf{OED Adaptive Gate ($M_{\mathrm{gate}}$)} & $88.00 \pm 2.03$ & $\mathbf{87.27 \pm 1.87}$ & $\mathbf{-0.73}$ & $\mathbf{88.77 \pm 1.89}$ \\
\midrule
\multirow{4}{*}{UCI Breast Cancer Stream}
 & Ungated SGD & $93.92 \pm 1.89$ & $15.77 \pm 4.70$ & $-78.15$ & $91.22 \pm 3.01$ \\
 & L2 Norm Clipping ($1.0$) & $93.88 \pm 1.98$ & $92.20 \pm 2.11$ & $-1.68$ & $95.14 \pm 1.84$ \\
 & Fixed Slow LR ($0.1\times$) & $89.79 \pm 2.33$ & $43.39 \pm 20.67$ & $-46.40$ & $89.41 \pm 2.56$ \\
 & \textbf{OED Adaptive Gate ($M_{\mathrm{gate}}$)} & $93.78 \pm 1.83$ & $\mathbf{93.43 \pm 1.63}$ & $\mathbf{-0.35}$ & $\mathbf{95.42 \pm 1.88}$ \\
\bottomrule
\end{tabular}%
}
\end{table*}

The resilience separation in Table~\ref{tab:exp3} is stark:
\begin{itemize}
    \item On \textbf{Two-Moons Stream}, Ungated SGD crashes by $-73.37$ pp (to $14.80\%$ mean trough), Slow-LR SGD crashes by $-66.37$ pp, and even L2 Norm Clipping collapses by $-45.40$ pp (to $42.77\%$) because unit-norm clipping still steps in the poisoned gradient direction. \textbf{OED Adaptive Gate ($M_{\mathrm{gate}}$)} limits mean max drop to just \textbf{$-0.73$ pp} ($87.27\%$ trough; paired Wilcoxon trough $p = 0.00009$ vs.\ Ungated, $p < 10^{-5}$ vs.\ Norm Clipping) and finishes highest at $88.77\%$ ($p = 0.00015$ vs.\ Slow-LR).
    \item On \textbf{UCI Breast Cancer Stream}, $M_{\mathrm{gate}}$ limits mean max shock drop to \textbf{$-0.35$ pp} ($93.43\%$ trough vs.\ $15.77\%$ Ungated, trough $p = 0.00009$; vs.\ $92.20\%$ Norm Clipping, trough $p = 0.0167$) and achieves the top final accuracy of $95.42\%$ ($p = 0.00009$ vs.\ Slow-LR).
\end{itemize}

\section{Conceptual Optical Equivalence and Scope Limitations}
\label{sec:discussion}

\subsection{Conceptual Analog Optical Mapping}
While digital edge processors execute Eq.~\eqref{eq:quad_map} iteratively in arithmetic loops, OED maps conceptually onto coherent analog optical computing \cite{wetzstein2020inference}: a spatial light modulator encodes the $24\text{-byte}$ seed onto a coherent wavefront, a $4f$ optical Fourier lens evaluates continuous spatial transforms at light-speed latency, boundary filters impose the cardioid cut-off ($c=1/4$), and photodetector arrays read non-divergent intensity into activation currents. Physical optical fabrication and hardware-level calibration remain future engineering milestones subject to Patent \texttt{TR 2026/016285}.

\subsection{Limitations and Future Research Directions}
We state three clear boundaries of this v4.0 preprint:
\begin{enumerate}
    \item \textbf{Single-Pass Capacity Bound:} As proved in Paper~1 \cite{dagli2026fns}, a single $24\text{-byte}$ seed carries at most $192$ bits of conditional Kolmogorov complexity; OED targets compact System-1 edge reflexes and continual streaming layers rather than post-hoc compression of billion-parameter language models.
    \item \textbf{Benchmark Scale:} The present preprint validates OED on 2D non-linear manifolds and moderate-dimensional ($d \le 30$) tabular diagnostic tasks; extending OED to high-dimensional continuous biosignals (building toward our whole-brain connectome work \cite{dagli2026werrsoma}), large-scale class-incremental continual learning, and equal-byte seed optimization versus seeded PRNGs remains active ongoing work.
    \item \textbf{Formal Dynamical Stability:} While algebraic invariants of OED are machine-checked in Lean~4 \cite{dagli2026lean4oed}, extending formal convergence and BIBO stability proofs to coupled continual-learning gradient-gating dynamics is reserved for future peer-reviewed journal extensions.
\end{enumerate}

\FloatBarrier
\section{Conclusion}
\label{sec:conclusion}

Orbital Error Dynamics (OED v4.0) reframes procedural neural synthesis as a non-equilibrium dynamical control problem along the complex quadratic boundary $\partial\mathcal{M}$. By unifying $R=256$ smooth boundary surfing, heavy-tailed Cauchy stochastic escape, and adaptive sigmoid gradient gating under a constant $24\text{-byte}$ persistent hidden seed, OED achieves $13.1\times$--$104.4\times$ model-weight Flash reduction across non-linear and diagnostic benchmarks, breaks zero-gradient interior traps with $33\times$ fewer perturbations than Gaussian noise ($p = 1.76 \times 10^{-5}$), and shields online streaming layers against catastrophic impulse shocks within $0.35$--$0.73$ pp (trough $p = 0.00009$). All code, data telemetry, and reproduction tests are released openly at \url{https://github.com/pCwOrM/fractal-neural-synthesis}.

\section*{Patent Priority Notice}
The procedural weight derivation architectures, homeostatic boundary control, heavy-tailed stochastic jump operator, adaptive sigmoid gradient gating mask, and conceptual zero-storage hardware co-processor frameworks disclosed in this work are subject to national patent application \textbf{TR 2026/016285}, filed on September 22, 2026 at the Turkish Patent and Trademark Office (T\"{U}RKPATENT) by Volkan Da\u{g}l\i\ (Inventors: Volkan Da\u{g}l\i, Dr.\ Zerrin Da\u{g}l\i, Da\u{g}han Da\u{g}l\i).

\bibliographystyle{IEEEtran}
\bibliography{references}

@article{dagli2026fns,
  author    = {Volkan Da{\u{g}}l{\i} and Zerrin Da{\u{g}}l{\i} and Da{\u{g}}han Da{\u{g}}l{\i}},
  title     = {Fractal Neural Synthesis: Constant-Storage Procedural Weight Derivation via Complex Quadratic Boundary Dynamics},
  journal   = {arXiv preprint arXiv:2611.32393 [cs.LG]},
  year      = {2026},
  note      = {{Version} 4.0; {CERN} Zenodo Concept DOI: 10.5281/zenodo.22774934 (v4.0: 10.5281/zenodo.24495977); Elsevier SSRN: 7659198},
  url       = {https://arxiv.org/abs/2611.32393}
}

@article{dagli2026lean4oed,
  author    = {Volkan Da{\u{g}}l{\i} and Zerrin Da{\u{g}}l{\i} and Da{\u{g}}han Da{\u{g}}l{\i}},
  title     = {Zero-Storage Procedural Neural Synthesis and Orbital Error Dynamics: Formal Machine-Checked Verification in {Lean} 4},
  journal   = {arXiv preprint arXiv:2609.33066},
  year      = {2026},
  note      = {{CERN} Zenodo Concept DOI: 10.5281/zenodo.22983889},
  url       = {https://arxiv.org/abs/2609.33066}
}

@article{dagli2026werrsoma,
  author    = {Volkan Da{\u{g}}l{\i} and Zerrin Da{\u{g}}l{\i} and Da{\u{g}}han Da{\u{g}}l{\i}},
  title     = {Bio-Synthetic Neuromorphic Interfacing: Whole-Brain {Drosophila} Connectome Coupling via Zero-Storage Procedural Reflexes},
  journal   = {CERN Zenodo Research Repository},
  year      = {2026},
  doi       = {10.5281/zenodo.22996625},
  url       = {https://doi.org/10.5281/zenodo.22996625}
}

@book{milnor2006dynamics,
  author    = {John Milnor},
  title     = {Dynamics in One Complex Variable},
  edition   = {3rd},
  publisher = {Princeton University Press},
  address   = {Princeton, NJ},
  year      = {2006}
}

@article{douady1984etude,
  author    = {Adrien Douady and John H. Hubbard},
  title     = {{\'E}tude dynamique des polyn{\^o}mes complexes},
  journal   = {Publications Math{\'e}matiques d'Orsay},
  volume    = {84-02},
  year      = {1984}
}

@book{schrodinger1944life,
  author    = {Erwin Schr{\"o}dinger},
  title     = {What is Life? The Physical Aspect of the Living Cell},
  publisher = {Cambridge University Press},
  year      = {1944}
}

@book{prigogine1977self,
  author    = {Gr{\'e}goire Nicolis and Ilya Prigogine},
  title     = {Self-Organization in Nonequilibrium Systems: From Dissipative Structures to Order through Fluctuations},
  publisher = {John Wiley \& Sons},
  address   = {New York},
  year      = {1977}
}

@article{friston2010free,
  author    = {Karl Friston},
  title     = {The free-energy principle: a unified brain theory?},
  journal   = {Nature Reviews Neuroscience},
  volume    = {11},
  number    = {2},
  pages     = {127--138},
  year      = {2010}
}

@article{bak1987self,
  author    = {Per Bak and Chao Tang and Kurt Wiesenfeld},
  title     = {Self-organized criticality: An explanation of the $1/f$ noise},
  journal   = {Physical Review Letters},
  volume    = {59},
  number    = {4},
  pages     = {381--384},
  year      = {1987}
}

@article{woodruff2016zinc,
  author    = {Francesca E. Duncan and Emily L. Que and Nan Zhang and Emily C. Feinberg and Thomas V. O'Halloran and Teresa K. Woodruff},
  title     = {The zinc spark is an inorganic signature of human egg activation},
  journal   = {Scientific Reports},
  volume    = {6},
  pages     = {24737},
  year      = {2016}
}

@inproceedings{jin2017escape,
  author    = {Chi Jin and Rong Ge and Praneeth Netrapalli and Sham M. Kakade and Michael I. Jordan},
  title     = {How to escape saddle points efficiently},
  booktitle = {Proceedings of the 34th International Conference on Machine Learning (ICML)},
  pages     = {1724--1732},
  year      = {2017},
  organization = {PMLR}
}

@inproceedings{horowitz2014computing,
  author    = {Mark Horowitz},
  title     = {1.1 {Computing's} energy problem (and what we can do about it)},
  booktitle = {IEEE International Solid-State Circuits Conference (ISSCC) Digest of Technical Papers},
  pages     = {10--14},
  year      = {2014},
  organization = {IEEE}
}

@article{shumailov2024model,
  author    = {Ilia Shumailov and Zakhar Shumaylov and Yiren Zhao and Nicolas Papernot and Ross Anderson and Yarin Gal},
  title     = {{AI} models collapse when trained on recursively generated data},
  journal   = {Nature},
  volume    = {631},
  pages     = {755--759},
  year      = {2024}
}

@article{furness2012enteric,
  author    = {John B. Furness},
  title     = {The enteric nervous system and neurogastroenterology},
  journal   = {Nature Reviews Gastroenterology \& Hepatology},
  volume    = {9},
  number    = {5},
  pages     = {286--294},
  year      = {2012}
}

@article{sakaguchi2008regulatory,
  author    = {Shimon Sakaguchi and Tomoyuki Yamaguchi and Takashi Nomura and Masahiro Ono},
  title     = {Regulatory {T} cells and immune tolerance},
  journal   = {Cell},
  volume    = {133},
  number    = {5},
  pages     = {775--787},
  year      = {2008}
}

@article{wetzstein2020inference,
  author    = {Gordon Wetzstein and Aydogan Ozcan and Sylvain Gigan and Shanhui Fan and Dirk Englund and Marin Solja{\v{c}}i{\'c} and Christian Denz and David A. B. Miller and Demetri Psaltis},
  title     = {Inference in artificial intelligence with deep optics and photonics},
  journal   = {Nature},
  volume    = {588},
  pages     = {39--47},
  year      = {2020}
}

@inproceedings{li2018measuring,
  author    = {Chunyuan Li and Heerad Farkhoor and Rosanne Liu and Jason Yosinski},
  title     = {Measuring the Intrinsic Dimension of Objective Landscapes},
  booktitle = {International Conference on Learning Representations (ICLR)},
  year      = {2018}
}

@inproceedings{ha2017hypernetworks,
  author    = {David Ha and Andrew M. Dai and Quoc V. Le},
  title     = {{HyperNetworks}},
  booktitle = {International Conference on Learning Representations (ICLR)},
  year      = {2017}
}

@inproceedings{chen2015compressing,
  author    = {Wenlin Chen and James T. Wilson and Stephen Tyree and Kilian Q. Weinberger and Yixin Chen},
  title     = {Compressing Neural Networks with the Hashing Trick},
  booktitle = {International Conference on Machine Learning (ICML)},
  pages     = {2285--2294},
  year      = {2015},
  organization = {PMLR}
}

@inproceedings{rahimi2007random,
  author    = {Ali Rahimi and Benjamin Recht},
  title     = {Random Features for Large-Scale Kernel Machines},
  booktitle = {Advances in Neural Information Processing Systems (NeurIPS)},
  volume    = {20},
  pages     = {1177--1184},
  year      = {2007}
}

@article{huang2006extreme,
  author    = {Guang-Bin Huang and Qin-Yu Zhu and Chee-Kheong Siew},
  title     = {Extreme learning machine: Theory and applications},
  journal   = {Neurocomputing},
  volume    = {70},
  number    = {1--3},
  pages     = {489--501},
  year      = {2006}
}

\end{document}